\documentclass{article}

\usepackage[nonatbib,final]{neurips_2024}
\usepackage[numbers]{natbib}

\usepackage[utf8]{inputenc} %
\usepackage[T1]{fontenc}    %
\usepackage{hyperref}       %
\usepackage{url}            %
\usepackage{booktabs}       %
\usepackage{amsfonts}       %
\usepackage{nicefrac}       %
\usepackage{microtype}      %
\usepackage[table]{xcolor}         %

\usepackage{enumitem}
\usepackage{array,multirow,pifont,xspace}
\usepackage{amsmath}
\usepackage{amssymb}
\usepackage{mathtools}
\usepackage{amsthm}
\usepackage{wrapfig}

\usepackage[capitalize,noabbrev]{cleveref}
\usepackage{adjustbox}

\renewcommand{\paragraph}[1]{\vspace{1.25mm}\noindent\textbf{#1}}
\newlength\savewidth\newcommand\shline{\noalign{\global\savewidth\arrayrulewidth
  \global\arrayrulewidth 1pt}\hline\noalign{\global\arrayrulewidth\savewidth}}

\newcommand{\tablestyle}[2]{\setlength{\tabcolsep}{#1}\renewcommand{\arraystretch}{#2}\centering\footnotesize}
\newcolumntype{x}[1]{>{\centering\arraybackslash}p{#1pt}}
\newcolumntype{y}[1]{>{\raggedright\arraybackslash}p{#1pt}}
\newcolumntype{z}[1]{>{\raggedleft\arraybackslash}p{#1pt}}
\definecolor{degray}{gray}{.6}
\newcommand{\deemph}[1]{\textcolor{degray}{#1}}
\newcommand{\app}{\raise.17ex\hbox{$\scriptstyle\sim$}}
\definecolor{baselinecolor}{gray}{.9}
\newcommand{\baseline}[1]{\cellcolor{baselinecolor}{#1}}

\def\acopy{Attention Copy\xspace}
\def\adistill{Attention Distillation\xspace}

\makeatletter
\DeclareRobustCommand\onedot{\futurelet\@let@token\@onedot}
\def\@onedot{\ifx\@let@token.\else.\null\fi\xspace}

\def\eg{\emph{e.g}\onedot} 
\def\ie{\emph{i.e}\onedot}

\def\wrt{w.r.t\onedot}

\makeatother

\title{On the Surprising Effectiveness of Attention Transfer for Vision Transformers}

\author{%
Alexander C. Li\thanks{Work done during an internship at FAIR.} \\
 Carnegie Mellon University \\
  \And
  Yuandong Tian \\
  FAIR \\
  \And
  Beidi Chen \\ %
  Carnegie Mellon University \\
  \And
  Deepak Pathak \\
  Carnegie Mellon University \\
  \And
  Xinlei Chen \\
  FAIR
}

\begin{document}

\maketitle

\vspace{-.5em}
\begin{abstract}
Conventional wisdom suggests that pre-training Vision Transformers (ViT) improves downstream performance by learning useful representations. Is this actually true?
We investigate this question and find that the features and representations learned during pre-training are not essential. 
Surprisingly, using only the attention patterns from pre-training (\ie, guiding how information flows between tokens) is sufficient for models to learn high quality features from scratch and achieve comparable downstream performance.
We show this by introducing a simple method called attention transfer, where only the attention patterns from a pre-trained teacher ViT are transferred to a student, either by copying or distilling the attention maps. 
Since attention transfer lets the student learn its own features, ensembling it with a fine-tuned teacher also further improves accuracy on ImageNet.
We systematically study various aspects of our findings on the sufficiency of attention maps, including distribution shift settings where they underperform fine-tuning.
We hope our exploration provides a better understanding of what pre-training accomplishes and leads to a useful alternative to the standard practice of fine-tuning. Code to reproduce our results is at \href{https://github.com/alexlioralexli/attention-transfer}{https://github.com/alexlioralexli/attention-transfer}.
\end{abstract}

\vspace{-.5em}
\section{Introduction\label{sec:intro}}

Pre-training has emerged as a dominant paradigm in machine learning and has significantly improved performance on a variety of tasks~\cite{howard2018universal,Devlin2019,Brown2020,he2022masked}. In computer vision in particular, self-supervised representation learning methods~\cite{He2020,Chen2020,Caron2021,he2022masked} and weakly supervised methods~\cite{Mahajan2018,Radford2021} have enabled learning from large amounts of images. It is widely accepted that these methods work because they teach models useful features that are relevant for downstream tasks. But is this story actually true? Perhaps there is another capability learned during pre-training that is sufficient to explain its benefits.

In this paper, we present an alternative explanation: pre-training teaches the model how information should be routed between tokens. We specifically focus on Vision Transformers (ViT)~\cite{dosovitskiy2020image}, not only because they are the most popular architecture for scaling, but also because Transformers explicitly \textit{decouple} this information flow. Inter-token communication is solely fulfilled by attention, while the remaining bulk of computation are intra-token operations that are applied to each token independently. In contrast, other architectures such as ConvNets~\cite{LeCun1989,He2016} simultaneously expand the receptive fields and extract the features, making it difficult to isolate the effect of information flow.  We hypothesize that the features computed by the intra-token operations are not essential to explain the benefits of pre-training, and that the pre-trained attention maps are typically sufficient for downstream tasks.

We test our hypothesis by introducing a new set of methods called attention transfer. Concretely, we treat a pre-trained ViT as the teacher and train a student model for downstream tasks while transferring only the attention patterns from the teacher. In contrast to the common fine-tuning paradigm of transferring all the weights (which mixes the effect of features and attention maps), \emph{only} the inter-token flow is transferred. In this way, the student must learn features from scratch, while isolating the benefits of the attention maps learned during pre-training.

\begin{wrapfigure}{r}{0.52\textwidth}
\centering
\vspace{-1em}
\includegraphics[width=\linewidth]{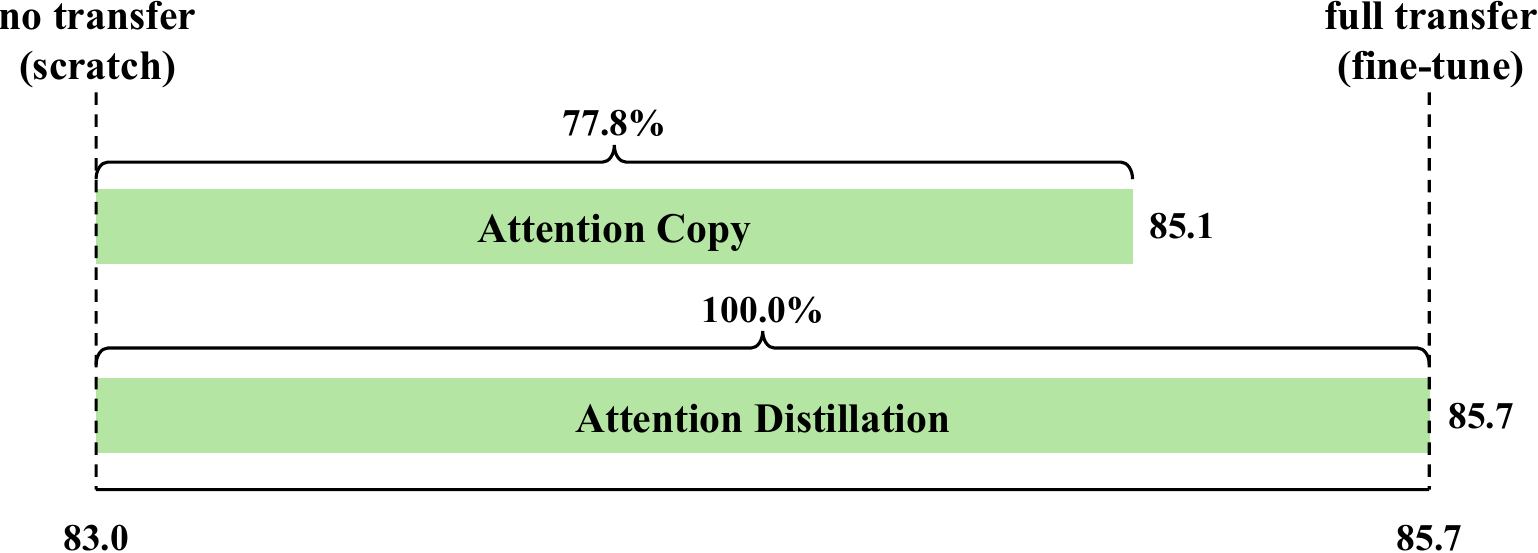}
\caption{\label{fig:teaser}\textbf{Using only attention is sufficient for full performance}. 
By copying the attention maps (top) from a MAE~\cite{he2022masked} pre-trained ViT-L~\cite{dosovitskiy2020image}, a ViT-L can reach a top-1 accuracy of 85.1 on ImageNet-1K~\cite{Deng2009} -- recovering 77.8\% of the gap between no transfer (training from scratch, 83.0) and full transfer (fine-tuning all the weights, 85.7). Distilling attention maps (bottom) can even \emph{fully} match MAE weight tuning while only transferring the inter-token flow.
}
\vspace{-.5em}
\end{wrapfigure}
We study two types of attention transfer. The first is \emph{\acopy}, which directly ``copy-and-pastes'' the attention maps. The learning is fully decoupled, as inter-token computation is entirely from the teacher, and the student only learns intra-token patterns routed by the teacher's attention maps. This is well-suited as a scientific probe, but is less practical since both networks need to be forwarded during the inference. The second is \emph{\adistill}, where the student simply distills attention patterns from the teacher, whose attention maps are no longer used after training. This is practical, but also helps identify the importance of the teacher's inter-token information flow.

While both attention transfer variants are straightforward, we find them \emph{highly effective}. \cref{fig:teaser} illustrates this with a ViT-L~\cite{dosovitskiy2020image} pre-trained using Masked Autoencoding (MAE)~\cite{he2022masked}. Compared to no transfer (training from scratch) and full transfer (fine-tuning all the MAE weights), \acopy can close most of the gap in performance, whereas \adistill can \emph{match} the fine-tuning accuracy on ImageNet-1K classification~\cite{Deng2009}. This is achieved by only transferring the inter-token flow from the same model. Furthermore, since attention transfer requires the student to learn features from scratch, those features are significantly different from the teachers' (\cref{fig:cka}) and improve ImageNet-1K accuracy score to 86.3 (+0.6) when ensembled with the teacher (\cref{fig:ensemble}).

To summarize, we make the following contributions:
\setlist{nosep}\vspace{.2em}
\begin{itemize}
    \item \textbf{Detailed analysis on the sufficiency of attention maps}. We find that solely using the pre-trained attention patterns is typically \textit{sufficient} to achieve the same downstream accuracy as fine-tuning on ImageNet-1K. Furthermore, we observe practical benefits, as ensembling with attention transfer significantly improves ImageNet performance. %
    This calls into question the commonly-believed story that pre-training is only about feature learning. While our main observation is robust \wrt different models and pre-training methods, we \textit{do find settings where pre-trained features are indeed necessary} to realize the full gains from pre-training. Our bare-minimum solution for attention transfer is more affected by data distribution shifts compared to weight tuning.
    \cref{sec:analysis} presents extensive analyses to better understand the behaviors of attention transfer. They are i) partial transfer with a subset of layers or heads; ii) variants of our method that transfer other attention-related activations; and importantly, iii) various ways to verify that the student is \emph{not} just re-learning the teacher model. \cref{sec:gen_and_limit} systematically tests how well our findings apply across a variety of pre-training and fine-tuning datasets, pre-training methods, model sizes, and tasks.
    \vspace{.2em}
    \item \textbf{Attention transfer methods}. We introduce \acopy and \adistill, which are methods to train a ViT on a downstream task while utilizing only the attention maps of a pre-trained teacher ViT. These methods help us understand the role of the features versus the attention patterns learned during pre-training. With further research, attention transfer could offer a potential alternative to the decade-long practice of fine-tuning pre-trained vision models~\cite{Girshick2014,dosovitskiy2020image,he2022masked}. Nearly all aspects of the fine-tuning pipeline have been thoroughly examined, suggesting a probable saturation of recipes. Weight sharing can also face security risks (\eg, white-box attacks~\cite{goodfellow2014explaining}). We hope our systematic examination of attention transfer sheds new light on how to leverage pre-trained ViTs, and will help establish this approach as an effective alternative when weight transfer is less applicable. 
\end{itemize}

\begin{figure*}[t]
    \centering
    \includegraphics[width=\linewidth]{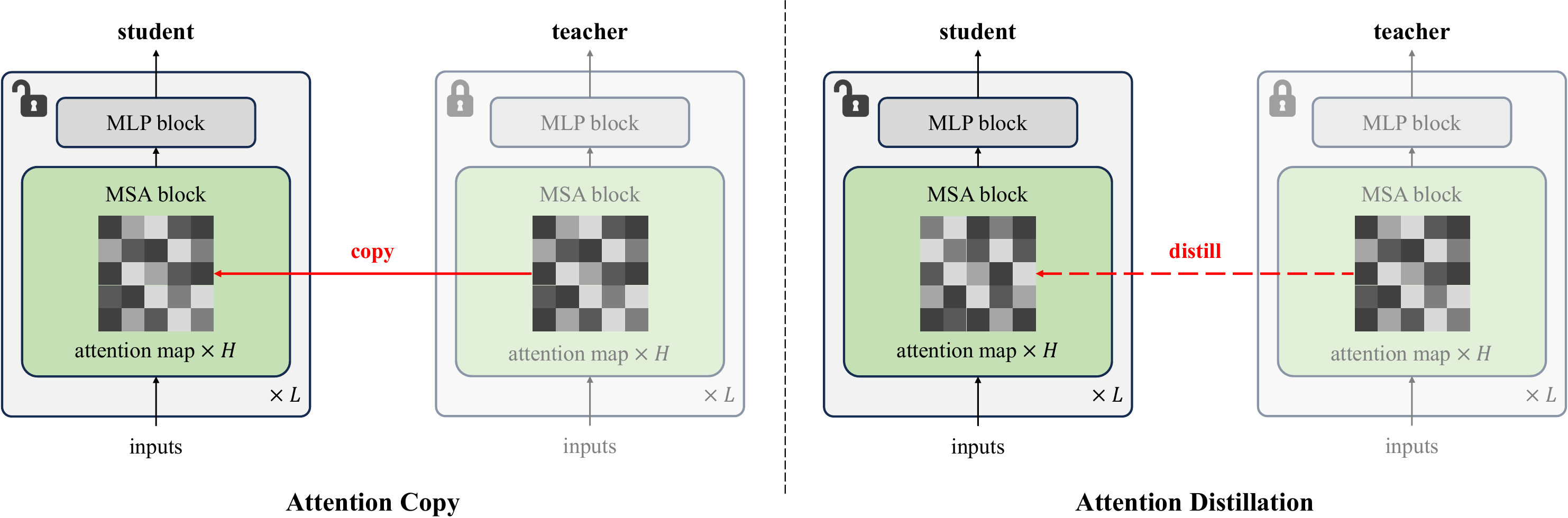}
    \vspace{-1.5em}
    \caption{\label{fig:method}Two types of \textbf{Attention transfer} for Vision Transformers. \textbf{Attention Copy} (left): We simply ``copy-and-paste'' the attention maps from a pre-trained teacher model to a randomly initialized student one. Other weights of the student are then trained via supervised learning. This fully decouples inter-token learning (from the teacher) and intra-token learning (in the student); but is less practical.
    \textbf{Attention Distillation} (right): The student computes its own attention maps, with an additional cross-entropy loss to distill patterns from the teacher during training. The teacher is no longer used during inference. $H$: number of heads; $L$: number of Transformer layers.}
    \vspace{-.5em}
\end{figure*}

\section{Attention Transfer\label{sec:transfer}}

\subsection{Attention Preliminaries\label{sec:prelim}}
To work with a Vision Transformer (ViT)~\cite{dosovitskiy2020image}, an image is first ``patchified'' into $N$ tokens. Their intermediate activations are represented as a sequence $X{=}\left[x_1, x_2, \cdots, x_N\right]^\top$, $x_i{\in}\mathbb{R}^C$, where $C$ is the embedding dimension.
The self-attention~\cite{vaswani2017attention} mechanism mainly introduces three learnable parameters $W_q, W_k, W_v{\in}\mathbb{R}^{C\times C/H}$ ($H$ is the number of heads). 
$Q{=}XW_q$ is often referred to as the queries, $K{=}XW_k$ as the keys, and $V{=}XW_v$ as the values. 
Then the attention function is defined as:
\begin{align}
\label{eq:attn}
    f_\text{attn} =\underbrace{\text{softmax}\left(QK^\top\right)}_{\text{attention map}}V,
\end{align}
where the softmax function is computed per query for the \emph{attention map}. Attention maps determine how the values from other tokens are aggregated, and with multiple heads, each token 
uses multiple attention distributions
within the same Multi-headed Self-Attention (MSA) block.

For an $L$-layer Transformer, MSA blocks are interleaved with MLP blocks, and each Transformer layer contains one of each block type. Most operations are \textit{intra-token computations}, which are applied independently to each token: value and projection matrices, normalization layers~\cite{ba2016layer}, and MLPs. The only \textit{inter-token computation} is applying the attention map $\text{softmax}(QK^\top)$, which is the only way for information to flow between tokens. Transformers are unique because their inter- and intra-token computations are \emph{decoupled}; however, the relative importance of each type of operation is not well understood, and Transformers are typically trained by \emph{jointly} fine-tuning all the weights.

Deviating from the common practice of joint weight tuning, we propose two attention transfer methods with the goal of exploring decoupled training for ViTs, described next.

\subsection{\acopy\label{sec:attn_copy}}
In this setup, we utilize two separate networks: a pre-trained teacher network that \emph{only} does a forward pass to compute its attention maps, and a student network that directly copies the attention maps from the teacher but computes all of the other activations. The student's weights are randomly initialized and trained via back-propagation, while the teacher's weights are kept frozen. This setting fully isolates the attention maps from the features that they are applied to, and thus is ideal for measuring the utility of pre-trained attention patterns when the student network learns to perform other tasks (\eg, image classification).

We call this method \textit{\acopy}, as we ``copy-and-paste'' the attention maps from teacher to student. \cref{fig:method} (left) shows a diagram of this approach. Note that it requires forward passes through both the teacher and student networks during the inference time. Given the extra computation, \acopy is not meant to be an entirely practical method. We mitigate this issue next.

\subsection{\adistill\label{sec:attn_distil}}

In \emph{\adistill}, the teacher network is only utilized at the training time. Given each training example, we forward both networks in parallel, with the student also computing its own attention maps. But besides the task-driven loss, we also enforce a distillation loss between student's attention maps and the teacher's counterparts as (soft) targets. Formally, using $Q_s {K_s}^\top$ for the student and $Q_t {K_t}^\top$ for the teacher, the loss is then defined as:
\begin{align}
    \mathcal{L}_\text{dist} = \mathcal{H}\left[\text{softmax}(Q_s{K_s}^\top), \text{softmax}(Q_t{K_t}^\top)\right],
\end{align}
where $\mathcal{H}$ computes the cross entropy.
As there can be multiple heads and layers in a Transformer, we simply sum up all the losses from wherever attention distillation is applied. Again, the student is trained via back-propagation. 
\cref{fig:method} (right) shows the diagram of \adistill.

Compared to \acopy, \adistill is much more practical. After training, the teacher is no longer needed, and the student can be used as a standalone model. Compared to training ViTs from scratch, the only addition is the distillation loss, meaning most of the optimization (\eg, learning rate, momentum) and regularization (\eg, weight decay, dropout rate~\cite{Srivastava2014}) hyperparameters can follow the scratch recipe with minimal adjustments. It does introduce a new hyperparameter $\lambda$, which weights the distillation loss and balances it with the task loss.

\adistill can be viewed as a form of generalized knowledge distillation, but it has several key differences from the design proposed by~\citet{hinton2015distilling}. \adistill trains the student to match the teacher's intermediate attention maps, not the final teacher output. This gives the flexibility of distilling from models trained on any task, not just models trained on the same final task. This property is well-suited for today's ``pre-train and transfer'' paradigm, where the pre-training task (\eg, reconstruction) and the downstream task (\eg, classification) are usually different.
However, \adistill does add the constraint that the architecture needs to compute attention maps. We leave experimenting on this idea for other architectures as future work.

Overall, while fancier designs can be used for both \acopy and \adistill, we choose to keep them simple for cleaner assessments of their effectiveness.

\subsection{Connection to Transformer Training Dynamics\label{sec:theory}}

Our investigation is also linked to recent attempts to theoretically understand the training dynamics of Transformers. Specifically, the inter-token flow encoded in the pre-trained attention maps can be regarded as a discovered latent \emph{hierarchy} from the dataset. Self-attention can quickly capture frequently co-occurring token pairs~\cite{jelassi2022vision,tian2023scan}. However, more occasional co-occurrences need to be explained by the top-level hierarchy, rather than directly learned in the lower levels~\cite{tian2023joma}. This is due to many potential spurious correlations~\cite{izmailov2022feature}, especially in the over-parameterized setting. Transferring attention maps from a trained teacher reduces these spurious inter-token correlations, so the student can focus on intra-token learning (\textit{i.e.}, computing useful features).

\section{Main Results\label{sec:main_results}}

As featured in \cref{fig:teaser}, attention transfer is highly effective despite its simplicity. Specifically, we demonstrate this with a ViT-L~\cite{dosovitskiy2020image} pre-trained with Masked Autoencoding (MAE)~\cite{he2022masked} for ImageNet-1K classification~\cite{Deng2009}. Note that this is the \emph{signature} result that established MAE's effectiveness for pre-training: compared to a ViT-L trained from scratch (with an accuracy score of 83.0), fine-tuning the MAE pre-trained on the same dataset results in a significant improvement to 85.7.\footnote{If not otherwise specified, our results are based on our faithful reimplementation of the official code in JAX.}

For attention transfer, we use the same pre-trained MAE model as our teacher, and since scratch training can be viewed as no transfer, and fine-tuning weights transfers all the weights, the above two results serve as natural lower and upper bounds for the effectiveness of our attention transfer methods. We make two observations (from \cref{tab:main_result} and \cref{fig:teaser}).

\subsection*{\emph{\acopy can largely close the gap between scratch training and full weight fine-tuning}}

We report an accuracy of 85.1 with \acopy. This has largely closed the gap between scratch and full weight tuning (to be precise, 77.8\% of the 2.7 percentage point gap). This is surprising, since the teacher's attention maps are frozen after pre-training for a different task (image reconstruction), and the student must learn everything else (the intra-token operations) completely from scratch.

As another upper-bound, we also experimented with \acopy from the \emph{fine-tuned} model. This reaches an accuracy score of 85.6 -- almost matching the teacher's performance (85.7), suggesting that adapting attention maps to the specific task is still helpful, but not crucial, especially as MAE pre-training is performed on the same data distribution. 

\subsection*{\emph{\adistill can match fine-tuning performance}}

Even more surprisingly, we find \adistill can achieve 85.7 -- \emph{on par} with fine-tuning the ViT-L weights from MAE. Since \adistill and weight tuning both result in the same-sized model, which requires the same compute budget for inference, this result suggests \adistill can be an effective drop-in replacement for weight tuning when the latter is less applicable (\eg, if weight sharing poses security risks, we can instead send the teacher's attention maps).

We hypothesize that distillation is better than copy because the student can choose how closely it matches the teacher attention maps, to better suit the task objective. This is also supported by the 85.6 accuracy of copying from fine-tuned MAE, which has the correct task-specific attention maps.

\begin{table}[t]
    \begin{minipage}[t]{0.48\textwidth}
    \centering
        \tablestyle{8pt}{1.2}
        \begin{tabular}{y{120}x{30}}
        method & acc. \\
        \shline
            scratch & 83.0 \\
            fine-tune & 85.7 \\
        \hline
            attn. copy & 85.1 \\ 
            attn. copy from fine-tuned & 85.6 \\
            attn. distill & 85.7 \\
        \hline
            ensemble attn. distill $+$ fine-tune & \textbf{86.3} \\
        \end{tabular}
        \vspace{0.5em}
        \caption{\label{tab:main_result}\textbf{Main results}. We show that the pre-trained attention patterns are \textit{sufficient} to match fine-tuning accuracy on ImageNet. Attention Copy closes most of the gap, and Attention Distillation achieves the same accuracy.
        }
    \end{minipage}
    \hfill
    \begin{minipage}[t]{0.48\textwidth}
    \centering
        \tablestyle{8pt}{1.2}
        \begin{tabular}{y{80}x{40}}
        transfer target & acc. \\
        \shline
        $Q$ & 85.6 \\
        $K$ & 84.4 \\
        $V$ & 84.4 \\
        \baseline{$Q$, $K$} & \baseline{85.1} \\
        \end{tabular}
        \vspace{0.5em}
        \caption{\label{tab:copy_subset}\textbf{Transfer other attention activations}. We test copying alternative attention activations other than the attention map -- $\text{softmax}(QK^\top)$. All alternatives do better than training from scratch, and transferring queries $Q$ actually does better than transferring the attention map.
        }
    \end{minipage}
    \vspace{-1em}
\end{table}

\section{Analysis\label{sec:analysis}}

Next, we provide extensive analyses to better understand the effectiveness of attention transfer. Broadly speaking, the explorations are driven by the following two questions:
\setlist{nosep}\vspace{1.25mm}
\begin{enumerate}[label=(\roman*)]
    \item How important are different activations, layers, and heads? (\cref{sec:variants}) %
    \item Is attention transfer re-learning everything from the teacher? (\cref{sec:relearn})
\end{enumerate}
\setlist{}

\subsection{Variants of Attention Transfer\label{sec:variants}}

We study four variants of attention transfer. We use \acopy within this section, since it is a fully-decoupled setting well-suited for scientific analysis. 

\begin{figure}[t]
    \begin{minipage}[b]{0.48\textwidth}
        \centering
        \includegraphics[width=\linewidth]{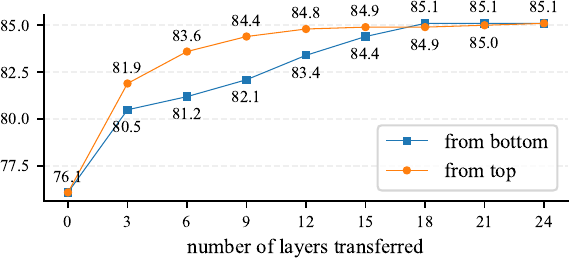}
        \caption{\label{fig:block}\textbf{Copy a subset of layers.} By default, all 24 ViT-L layers are transferred. Here we only transfer a subset, and find: more layers always helps; and attention maps from top layers are more beneficial than those from bottom layers.}

    \end{minipage}
    \hfill
    \begin{minipage}[b]{0.48\textwidth}
        \centering
        \includegraphics[width=\linewidth]{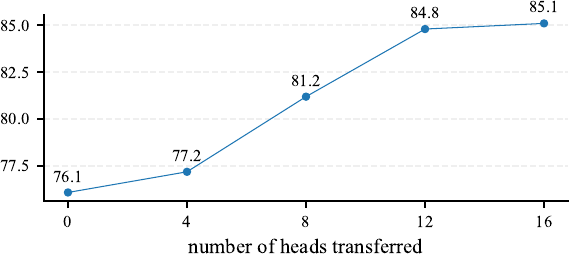}
        \caption{\label{fig:head}\textbf{Copy a subset of heads.} The pre-trained ViT-L has 16 heads in each MSA block. By default, all of them are transferred. Here we only transfer a subset, and find more heads helps in general, but performance saturates at 12 heads.}
    \end{minipage}
    \vspace{-1em}
\end{figure}

\paragraph{Transfer a subset of $Q$, $K$ and $V$.}
A natural alternative to transferring attention maps is to transfer different activations that come with self-attention (Eq.~\ref{eq:attn}), namely queries $Q$, keys $K$, or values $V$. Without loss of generality, if we transfer the teacher's $Q$, the student will compute its own $K$ and $V$ and use them normally. Note that transferring both $Q$ and $K$ is equivalent to transferring the map $\text{softmax}(QK^\top)$. \cref{tab:copy_subset} shows that transferring $Q$ works surprisingly well, and is actually better than transferring the attention map.

We suggest that copying $Q$ gives the model the flexibility to deviate from the teacher attention maps and use attention patterns that are better suited for the downstream task. This is supported by the fact that copying $Q$ and \acopy from the fine-tuned model both achieve the same accuracy of $85.6$. \cref{sec:head_jsd} dives deeper into this hypothesis and finds that the attention maps for copying $Q$ are similar to the teacher's but less constrained than they are in other transfer methods. While more investigation could be done in future work, our findings suggests that the queries $Q$ are more important than the keys, which is consistent with previous findings in text sequence modeling where the number of keys and values per layer can be significantly reduced~\cite{shazeer2019fast}.

\begin{wraptable}{r}{0.3\textwidth} %
\centering
        \vspace{-1.5em}
        \tablestyle{8pt}{1.2}
        \begin{tabular}{y{40}x{30}}
        method & acc. \\
        \shline
            attn. distill & 85.7 \\
            $Q$ distill & 81.3 \\
        \end{tabular}
    \caption{Attention Distillation outperforms $Q$ distillation.}
    \label{tab:q_distill}
    \vspace{-1em}
\end{wraptable}

Finally, we test whether distilling $Q$ could outperform full \adistill. However, \cref{tab:q_distill} shows that $Q$ distillation does significantly worse. We hypothesize that this is because it is harder for the student to learn useful keys $K$ while $Q$ is still being learned, and because \adistill already has the flexibility to adjust its attention maps to suit the downstream task.

\paragraph{Partial transfer: layers.}
We next change the number of Transformer layers transferred, aiming to identify which layers are more valuable from the teacher. The baseline transfers all the layers. In \cref{fig:block}, we try
transferring attention maps only from the first or last layers. For the remaining layers, the student learns to compute its own attention maps.

We make several observations: i) We find transferring more layers is always more helpful. This is a bit surprising, as one may expect attention patterns optimized for MAE's patch-wise reconstruction task to be sub-optimal for a high-level recognition task like classification. ii) We find transferring the later attention maps is generally better. In fact, performance roughly saturates when transferring the last 15 attention maps out of a total of 24. This indicates that ViTs are capable of learning good low-level representations, as long as they are told how to combine these features into higher-level ones; but not vice versa. This reinforces the theory from~\citet{tian2023joma} that guidance on top-level hierarchy is more important, as there are many more possibilities, and attention transfer can reduce possible spurious correlations in the lower levels.

\paragraph{Partial transfer: heads.}
Finally, we switch back to transferring all the layers, but change the number of heads copied from each MSA block. Specifically, instead of copying the attention map from every head, we can selectively choose to use a subset of the teacher's heads. The student can then compute its own attention patterns for the unused heads.
\cref{fig:head} shows the effect of transferring fewer heads for each layer. Performance improves as we do attention transfer with more heads, though performance almost saturates at 12 out of 16 heads. Note that we simply follow a na\"{i}ve selection strategy and use the first set of heads; more advanced strategies based on diversity or activation magnitude can potentially improve the robustness as we reduce the number of heads.

\subsection{Are We Re-Learning the Teacher?\label{sec:relearn}} 
Since attention transfer provides a significant amount of information from the teacher (ViT-L attention maps have about 10M activations total per image, see~\cref{sec:bits} for detailed calculations), a natural question is whether the student performs well because it simply relearns the same representations as the teacher. We thoroughly test this hypothesis on different aspects of the student model.

\begin{figure}[t]
    \centering
    \begin{minipage}{0.48\linewidth}
        \includegraphics[width=\linewidth]{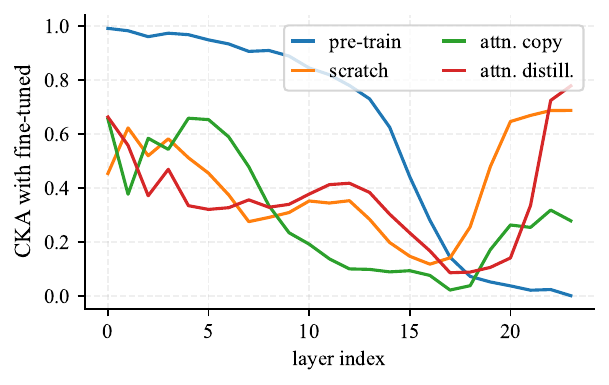}
    \end{minipage}
    \hfill
    \begin{minipage}{0.48\linewidth}
        \caption{\label{fig:cka}\textbf{CKA representation similarity to the fine-tuned model.} We use CKA~\cite{kornblith2019similarity} to measure the layer-wise similarity between representations learned in different models against the fine-tuned MAE model. Higher means more similar. We find that attention transfer methods are quite \emph{dissimlar} to the fine-tuned model, with roughly the same CKA as an independent scratch model.}
    \end{minipage}
    \vspace{-1.5em}
\end{figure}

\paragraph{Representation similarity.}
One way that the student can reproduce the teacher is by learning the same intermediate features. We measure this using Centered Kernel Alignment (CKA)~\cite{kornblith2019similarity}, a similarity measure for representations that has been shown to be effective even for networks trained with different initializations or architectures. CKA is a layer-wise comparison that ranges from 0 (completely dissimilar) to 1 (identical) and is invariant to orthogonal linear transformations and isotropic scaling of the features. \cref{fig:cka} shows the CKA between our fine-tuned model and our attention transfer methods. We also show the pre-trained model and a ViT-L trained from scratch for reference. We compute CKA with respect to the fine-tuned model, even though we copy or distill from the pre-trained MAE model, since the features change significantly during fine-tuning to become more task-specific. 
Overall, \acopy and \adistill are both quite \textit{dissimilar} to the fine-tuned MAE model, following the same similarity trend as the scratch model. Our sanity check passes, as CKA shows that pre-trained and fine-tuned MAE have very similar representations in the early layers. This is expected since fine-tuning with layer-wise learning rate decay~\cite{Clark2020} means the earliest layers change very little during fine-tuning. 

\begin{figure}[t]
    \centering
    \begin{minipage}{0.5\linewidth}
        \includegraphics[width=\linewidth]{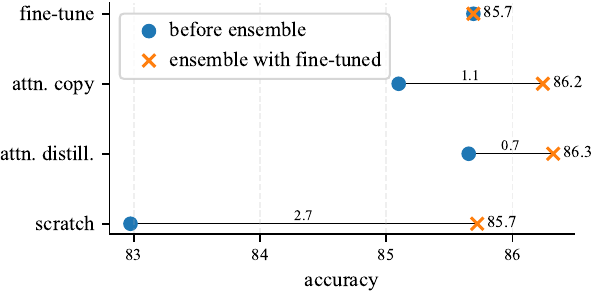}
    \end{minipage}
    \hfill
    \begin{minipage}{0.48\linewidth}
        \caption{\textbf{Ensemble accuracy with the fine-tuned model}. We plot the accuracy of ensembling our attention transfer models and a fine-tuned MAE. We use this to measure model prediction similarity with the fine-tuned model. The ensemble yields notable accuracy gains over the base model, reaching as much as 86.3, indicating that the attention transfer based models are \emph{not} particularly correlated with the fine-tuned model.}
        \label{fig:ensemble}
    \end{minipage}
\end{figure}

\paragraph{Prediction similarity and ensembling.} Our CKA analysis may not catch some similarity of the intermediate representations, as CKA does not detect all forms of the same information (\eg, invertible nonlinear transforms)~\cite{kornblith2019similarity}. Since intermediate representations may not tell the full story, we also examine the similarity of the network outputs. We measure this using network ensembling: given softmax predictions $p_\text{ft}$ from the fine-tuned model and $p_\text{other}$ from another model, we test the accuracy of their average $(p_\text{ft} + p_\text{other})/2$. 
The more independent the model predictions are, the higher their ensemble accuracy is. Figure~\ref{fig:ensemble} compares accuracy before and after ensembling with the fine-tuned model. 
Attention transfer is dissimilar enough to achieve high ensemble accuracy, and ensembling \adistill with a fine-tuned MAE achieves 86.3, +0.6 over the fine-tuned MAE model.

Finally, \cref{sec:attn_map_vis} visualizes the attention maps learned by \adistill and shows that they match for distilled attention blocks but are drastically different for layers that are not distilled. 

\section{Generalization and Limitations\label{sec:gen_and_limit}}
In this section, we test how well our findings on attention transfer apply across a variety of pre-training and fine-tuning datasets, pre-training methods, model sizes, and tasks.

\begin{table}[t]
    \begin{minipage}[t]{0.48\textwidth}
    \centering
        \tablestyle{8pt}{1.2}
        \begin{tabular}{y{80}x{15}x{15}x{40}}
        pre-training data & tune & copy \\
        \shline
            ImageNet & 85.7 & 85.1 \\
            ImageNet-22K & 85.5 & 84.4 \\
            COCO & 85.2 & 83.1 \\
        \end{tabular}
        \vspace{0.5em}
        \caption{\label{tab:laion}\textbf{Different pre-training datasets}. We pre-train MAE on more datasets, and then either fine-tune or do copy for ImageNet-1K classification. Attention transfer works well when the data distribution stays stable, but its effectiveness is more negatively affected by distribution shifts.
        }
    \end{minipage}
    \hfill
    \begin{minipage}[t]{0.48\textwidth}
    \centering
        \tablestyle{8pt}{1.2}
        \begin{tabular}{y{35}x{15}x{15}x{15}x{25}}
            eval. data & scratch &  tune & copy & distill \\
            \shline
            iNat 2017 & 49.7 & 75.9 & 69.1 & 69.3 \\
            iNat 2018 & 64.3 & 79.9 & 71.8 & 74.1 \\
            iNat 2019 & 66.2 & 83.8 & 77.9 & 80.0 \\
        \end{tabular}
        \vspace{0.5em}
        \caption{\textbf{Long-tail recognition on iNaturalist}, with ImageNet-1K pre-trained MAE. We tune weights or do attention transfer on iNaturalist, and we again find attention transfer is worse than tuning weights when the pre-training dataset is different from the downstream dataset.}
        \label{tab:dataset}
    \end{minipage}
    \vspace{-0.5em}
\end{table}

\subsection{Pre-training and fine-tuning datasets}
So far, we have focused on a MAE model pre-trained and evaluated on ImageNet-1K. What happens if we pre-train or evaluate on different datasets? We first test this by pre-training MAE ViT-L models on two new datasets: ImageNet-22K~\cite{Deng2009} and COCO~\cite{Lin2014}. These have substantially different dataset bias~\cite{torralba2011unbiased} from ImageNet, across axes like appearance, class balance, and diversity. \cref{tab:laion} shows that the resulting MAE models maintain relatively good performance when fine-tuning on ImageNet-1K, with a maximum drop of at most 0.5. However, Attention Copy accuracy drops more, losing as much as 2.1.  We see a similar phenomenon in \cref{tab:dataset} where we use a MAE pre-trained on ImageNet and transfer to the iNaturalist datasets~\cite{VanHorn2018}. Again, when the pre-training dataset does not match the transfer dataset, Attention Copy accuracy drops significantly. We hypothesize that the frozen teacher's attention maps are ill-suited for the fine-tuning dataset, which limits the performance.

\subsection{Out-of-distribution robustness}
One notable aspect of a standard fine-tuned MAE model is that it shows slight ``effective robustness,'' \textit{i.e.}, it achieves slightly better out-of-distribution (OOD) accuracy than expected based on its in-distribution (ID) accuracy~\cite{fang2022data}. We test whether Attention Distillation, which achieves the same ID accuracy, has the same benefits OOD. \cref{tab:robustness} shows that Attention Distillation still does quite well, but has lower accuracy than fine-tuned MAE on all 4 distribution shifts we tried. These results indicate that the attention maps do not account for the full robustness benefits, and that the features learned by MAE during pre-training are helpful OOD even if they are not ID. 

\begin{table}[t]
\centering
\tablestyle{8pt}{1.2}
\begin{tabular}{y{110}x{15}x{15}x{15}x{25}}
    out-of-distribution evaluation & scratch & tune & copy & distill\\
    \shline
    ImageNet-A~\cite{Hendrycks2021}      & 32.0 & 56.5 & 48.9 & 54.3 \\
    ImageNet-R~\cite{Hendrycks2021a}     & 51.9 & 59.6 & 57.5 & 56.8 \\
    ImageNet-S~\cite{Wang2019}           & 38.0 & 45.2 & 43.1 & 42.9 \\
    ImageNet-V2~\cite{recht2019imagenet} & 72.4 & 76.4 & 75.5 & 75.9 \\
\end{tabular}
\vspace{0.5em}
\caption{\textbf{Out-of-distribution robustness}. We take two models that achieve the same accuracy on ImageNet-1K (fine-tuned and distilled), and evaluate them on a suite of distribution shifts. Attention Distillation does well when the distribution is close (\eg, on ImageNet-V2), but loses the mild ``effective robustness'' that fine-tuned MAE has been found to have~\cite{fang2022data}.}
\label{tab:robustness}
\vspace{-.5em}
\end{table}

\begin{table}[t]
    \begin{minipage}[t]{0.48\textwidth}
    \centering
        \tablestyle{8pt}{1.2}
        \begin{tabular}{y{75}x{15}x{15}x{15}}
            pre-training method & tune & copy & distill \\
            \shline
            MAE~\cite{he2022masked} & 85.7 & 85.1 & 85.7 \\
            MoCo-v3~\cite{Chen2021a}    &  84.0 & 82.5 & 83.3\\
            FLIP~\cite{li2023scaling}   & 87.4 & 86.6 & 86.1\\
            DINO\textsuperscript{\textdagger}~\cite{Caron2021} & 83.2 & 82.3 & 82.8 \\
            \deemph{none}               & \deemph{83.0} & \deemph{72.7} & \deemph{76.3} \\
        \end{tabular}
        \vspace{0.5em}
        \caption{\textbf{Different pre-training methods.} Attention transfer works for all pre-training methods, even reaching 86.6 with a FLIP teacher. As a sanity check, transferring from a randomly initialized ViT significantly hurts. \textsuperscript{\textdagger}DINO is ViT-B.
        }
        \label{tab:init}
    \end{minipage}
    \hfill
    \begin{minipage}[t]{0.48\textwidth}
    \centering
        \tablestyle{8pt}{1.2}
        \begin{tabular}{y{26}x{20}x{20}x{20}x{20}}
         model    & scratch & tune & copy & distill\\
        \shline
            ViT-B & 82.5 & 83.6 & 82.0 & 83.4 \\
            ViT-L & 83.0 & 85.7 & 85.1 & 85.7 \\
            ViT-H & 83.0 & 86.9 & 86.1 & 86.3
        \end{tabular}
        \vspace{0.5em}
        \caption{\textbf{Different model size} with MAE pre-trained on ImageNet-1K. Similar to weight tuning, the classification accuracy of attention transfer scales well as we vary the model size, while scratch training saturates.%
        }
        \label{tab:model_scale}
    \end{minipage}
\end{table}

\begin{table}[t]
\centering
\tablestyle{8pt}{1.2}
\begin{tabular}{y{40}x{30}x{40}x{40}}
   metric & scratch & tune & distill \\
\shline
    $AP^{box}$ & 39.1 & 46.3 (+7.2)& 43.6 (+4.5) \\
    $AP^{mask}$ & 34.6 & 40.6 (+6.0) & 38.7 (+4.1) \\
\end{tabular}
\vspace{0.5em}
\caption{\textbf{Object detection results} on COCO with a MAE ViT-B pre-trained on COCO. Attention transfer achieves a majority of the gains of pre-training in this setting as well.}
\label{tab:detection}
\vspace{-2em}
\end{table}

\subsection{Pre-training methods}
We have so far focused on MAE, a reconstruction-based pre-training method. We now check whether attention transfer still works if the teacher has been pre-trained with a different algorithm. Specifically, we test MoCo-v3~\cite{Chen2021a}, a self-supervised contrastive learning approach, and FLIP~\cite{li2023scaling}, which does image-text contrastive learning. \cref{tab:init} shows that Attention Copy still achieves most of the performance benefits for each pre-training method. \textit{Impressively, ViT-L is even able to reach 86.6 by just transferring attention maps from FLIP}. This confirms that learning the proper attention patterns is indeed a significant bottleneck during learning. Note that the FLIP model we used is pre-trained on LAION-2B~\cite{schuhmann2022laion}, yet its effectiveness is less affected by distribution shifts to ImageNet-1K.

\subsection{Model size} 
We test whether attention transfer works across model sizes. For all experiments so far, we have used ViT-L; here, we try Attention Copy from a smaller (ViT-B) and larger (ViT-H) model, both pre-trained with MAE. 
\cref{tab:model_scale} shows that Attention Copy continues to improve with scale, even reaching 86.1\% accuracy with ViT-H. It can do this even though scratch model performance already saturates at the ViT-L size. Again, this indicates that models need proper inter-token routing in order to learn good features that generalize. Otherwise, they cannot properly utilize increased model capacity.

\subsection{Object Detection}
Finally, we examine the performance of attention transfer in the standard ViTDet pipeline~\cite{li2022exploring} for COCO object detection. We compare training from scratch against fine-tuning and attention transfer from a MAE ViT-B pre-trained on COCO, which is done to mitigate the effect of distribution shift. For fair comparisons, we use a $448 {\times}448$ input to remove the effect from window attention and positional embedding interpolation, and remove the effect of relative positional embeddings. \cref{tab:detection} shows that \adistill recovers a majority of the gains from pre-training in this dense prediction setting as well. Based on \cref{tab:model_scale}, we anticipate that the gap between fine-tuning and attention transfer will decrease with ViT-L, but we are limited by computational resources.

\section{Related Work\label{sec:related}}
\paragraph{Structure in attention patterns.}
Numerous previous works have studied the attention patterns of pre-trained vision transformers~\cite{walmer2023teaching,xie2023revealing,park2023self}. 
These works present these differences only as qualitative observations, whereas we are able to isolate the attention patterns and show that they are causally responsible for most of the differences in fine-tuning performance. Other methods, such as Lora~\cite{hu2021lora} or Prompt-to-Prompt~\cite{hertz2022prompt}, do rely on the importance of high quality attention patterns within pre-trained networks, but they also utilize pre-trained features and do not provide our insight that \textit{these features are typically unnecessary} for the tasks we examine.
\citet{trockman2023mimetic} observe diagonal structure within the product of attention layer weights in a trained supervised network. They show that initializing the weights with this structure moderately improves accuracy for small models early in training. 
The work most similar to us is \citet{zhang2022unveiling} in the language domain, which finds that pre-trained BERT models improve length generalization on a few particular synthetic tasks. They attribute it to the attention patterns of a few, specific heads and show that hardcoding these patterns into the network achieves the same benefit. Our work is complementary and emphasizes the importance of attention maps over features.

\paragraph{Decoupling inter- and intra-token operations.}
GLoMo~\cite{yang2018glomo} also attempts to decouple features from the way they should be combined. They use unsupervised pre-training to train a network to output a graph, which is later used to combine task-specific features. 
We find that there is no need to develop a specialized architecture to achieve this --  Vision Transformers \textit{already do this naturally}. 

\paragraph{Knowledge Distillation}
Knowledge distillation is a popular framework for training small, high-performing student networks~\cite{hinton2015distilling}. Knowledge distillation methods typically add a loss to encourage the student network to match the teacher network's logits, but variants often use other feature statistics as targets, such as the final representations~\cite{navaneet2022simreg,duval2023simple}, intermediate feature similarities~\cite{hao2022learning}, or intermediate feature magnitudes~\cite{zagoruyko2016paying,li2021neural}. This last approach has also previously been called ``attention transfer,'' but their method is quite different and actually refers to distilling spatial activation magnitudes in ConvNets. 
These knowledge distillation approaches all assume that students need to be explicitly taught the right features.
In contrast, our analysis with attention transfer shows that attention maps are sufficient to recover all of the gains from pre-training. 
Some papers have used attention distillation as an auxiliary loss to help a smaller model learn the teacher outputs more effectively~\cite{wang2020minilm,wang2022attention}. However, these only consider transferring the same function across model sizes, instead of transferring knowledge from a pre-trained model to a different downstream task.

\paragraph{Connection to the lottery ticket hypothesis}
 The lottery ticket hypothesis~\cite{frankle2018lottery} suggests that large, dense neural networks contain small, sparse subnetworks (``winning tickets'') that, when trained from scratch, can match or even outperform the performance of the original dense network. This is particularly interesting because these sparse subnetworks maintain their performance only with their original initialization values; the strength of their connections between \textit{neurons} is special in some way. Our findings draw a parallel, indicating that the connections between \textit{patches}, controlled solely by the attention patterns, are similarly special within pretrained ViTs. \citet{frankle2018lottery} further conjecture that overparameterization improves performance because larger models contain exponentially more sparse subnetworks in superposition and are thus more likely to contain a ``winning ticket'' -- a hypothesis supported by subsequent empirical and theoretical work \citep{ramanujan2020s,pensia2020optimal,orseau2020logarithmic,burkholz2022most}. However, this phenomenon does not arise in our setting with ViT attention patterns, since there are only a handful of attention maps per layer (rather than thousands of neurons). Consequently, good attention patterns are unlikely to appear by chance and must instead be learned during pretraining. We hope that a new model architecture that efficiently combines many more attention maps per layer can address this limitation and learn better from scratch than existing ViTs.

\section{Conclusion}
Even as Transformers have surged in popularity, the way we use them has remained stagnant: pre-train, then fine-tune the weights. 
In this work, we present attention transfer, a simple alternative to ViT fine-tuning that decouples intra-token operations (how to extract more usable features for each token) from inter-token operations (how those features should be combined).
Our key finding is that the attention patterns (inter-token operations) are the key factor behind much of the effectiveness of pre-training -- our Attention Distillation method \textit{completely matches fine-tuning} on ImageNet-1K.
We do find some limitations: attention transfer does not work well if the pre-training and transfer datasets are different, and it loses a bit of OOD robustness. Nevertheless, our findings provide insights into the role of attention in pre-trained ViTs, and we hope 
future work fixes attention transfer's shortcomings and explores the advantages of this new transfer method. 

Some directions for future work are particularly interesting. First, a deeper investigation of the queries $Q$ could help us better understand their importance and potentially yield better transfer strategies. 
Second, attention transfer eliminates the need for tricks that fine-tuning requires, such as layerwise learning rate decay. Layerwise learning rate decay adds the prior that early layers should change less compared to later layers. However, this prior may be overly restrictive for next-generation models, since it prevents early features from changing, and getting rid of it could open up new opportunities. Finally, since attention maps are $L \times L$, where $L$ is the sequence length, attention maps could be transferred more easily across model sizes. In contrast, weight tuning is more difficult to apply when the models have different dimensions. Pre-training a smaller model and transferring its attention patterns to a larger downstream model could be more practical than the current practice of fine-tuning.

\begin{ack}
AL is supported by the NSF GRFP DGE1745016 and DGE2140739 and performed the work during an internship at FAIR.
\end{ack}

\bibliography{at}
\bibliographystyle{abbrvnat}

\newpage
\appendix

\section*{Appendix}
\section{Key Numbers}
\subsection{Information in Attention Transfer\label{sec:bits}}
How much information is transferred during attention transfer? \cref{tab:params} shows two ways of doing this accounting. If considering the map as 24 layers $\times$ 16 heads $\times$ 197 query tokens $\times$ 197 key tokens, there are about 15 million activations transferred per example. However, $QK^\top$ is low rank since $Q$ and $K$ are very ``tall,'' so the attention map $\text{softmax}(QK^\top)$ can be considered 24 layers $\times$ 16 heads $\times$ 197 tokens $\times$ 64 head dim $\times$ 2 matrices, which is about 9.7 million activations.  
\begin{table}[h]
    \centering
    \begin{tabular}{lc}
    Accounting Method & Parameters \\
    \midrule
        Count sizes of $Q, K$ & $24 \times 16 \times 197 \times 64 \times 2 \approx$ 9.7M \\
        Count size of $QK^\top$ & $24 \times 16 \times 197 \times 197 \approx$ 15M \\
    \end{tabular}
    \vspace{0.5em}
    \caption{Number of parameters activations transferred per example for a ViT-L teacher. }
    \label{tab:params}
\end{table}

\subsection{Computational Cost of Attention Transfer}
\label{sec:cost}

Attention transfer has the same computational and memory cost as any other knowledge distillation method that does a forward pass through a teacher. We compare fine-tuning vs attention distillation on a 16GB NVIDIA GP100 with ViT-L and a batch size of 16:

\begin{table}[h]
    \centering
    \begin{tabular}{lcc}
    Method & Memory (GB) & Time per iteration (s) \\
    \midrule
        Weight Fine-tuning & 9.4 & 0.93 \\
        Knowledge Distillation & 11.1 & 1.23 \\
        Attention Copy/Distillation & 11.1 & 1.23 \\
    \end{tabular}
    \vspace{0.5em}
    \caption{Training cost of Attention Transfer. }
    \label{tab:cost}
\end{table}

Training these large models on ImageNet-1K is quite computationally expensive. 100 epochs of regular fine-tuning is about 2070 GPU-hours per 100 epochs, and 100 epochs of attention transfer is about 2735 GPU-hours. In total, we estimate about 150k GPU-hours are required to reproduce all  experiments.

\section{Additional Analysis}

\subsection{Aggregated Attention Transfer}
\begin{table}[t]
\centering
\tablestyle{8pt}{1.2}
\begin{tabular}{y{80}x{40}}
average over & acc. \\
\shline
examples & 79.7 \\
layers & 82.7 \\
heads & 82.2 \\
query tokens & 79.9 \\
\baseline{none} & \baseline{85.1} \\
\end{tabular}
\vspace{0.5em}
\caption{\label{tab:averaged}\textbf{Aggregated attention transfer}. The full set of attention maps from the teacher has shape (examples, layers, heads, query tokens, key tokens). Here we try to average over each of these axes \emph{before} the transfer. Performance drops the most when averaging over examples or query tokens, indicating that these are the most important aspects of the attention maps.
}
\end{table}
In \cref{sec:variants}, we conducted a thorough analysis of which aspects of attention matter the most. 
Another way to identify key properties of the teacher's attention maps is to average them across some axis during the transfer. For example, one can average the attention maps over all layers of the teacher network, so that the student uses the same map at every layer. \cref{tab:averaged} shows the results with aggregations over several natural axes. Averaging over examples (\ie, using the same attention map, independent of the input) or averaging over query tokens (\ie, each attention distribution is the same, regardless of the query token given an image) does quite poorly. This indicates, unsurprisingly, that these are key elements of self-attention. This also shows that prior work that focuses on aggregate statistics of the attention maps (\eg, averaged over examples)~\cite{walmer2023teaching} fail to capture the per-example nature of the attention maps that are actually responsible for full fine-tuning performance.
Attention copy performance is more reasonable when averaging over heads or layers. This partially corroborates previous findings that attention maps can largely be shared across all layers~\cite{venkataramanan2023skip}. However, while the results can be potentially improved with more recipe search, the performance is far short of the full fine-tuning accuracy (85.7). 

\subsection{Comparison to Knowledge Distillation}

\begin{table}[t]
\centering
\tablestyle{8pt}{1.2}
\begin{tabular}{y{80}x{40}}
method & acc. \\
\shline
attn. distill & \textbf{85.7} \\
feature distill & 81.3 \\
fine-tune & \textbf{85.7} \\
\baseline{scratch} & \baseline{79.7} \\
\end{tabular}
\vspace{0.5em}
\caption{\label{tab:kd}\textbf{Comparison with knowledge distillation}. We try knowledge distillation from the pre-trained teacher by adding an auxiliary MSE loss on the residual stream output. This encourages the student model to match the representations of the teacher (``feature distill''). We find that this does much worse than \adistill.
}
\end{table}

Our central hypothesis has been that the pre-trained attention maps are \textit{sufficient}, and the pre-trained features are not \textit{necessary}. Since our attention transfer methods are special instances of knowledge distillation~\cite{hinton2015distilling}, we additionally compare to a baseline of distilling the residual stream features from a pre-trained MAE ViT-L. In \cref{tab:kd}, we obtain a downstream accuracy of 81.3 on ImageNet-1k. This is significantly lower than the 85.7 that can be achieved through fine-tuning or attention distillation. This makes sense: the features learned during self-supervised pre-training are not directly well-suited for classification, so trying to match them can hurt performance. CKA analysis of the features (\cref{fig:cka}) supports this hypothesis – the fine-tuned MAE does well by significantly changing the features in the latter half of the network. Overall, transferring attention appears to do much better than distilling the features.

\subsection{Attention Map Analysis for Transferring $Q$}
\label{sec:head_jsd}
\begin{figure}
    \centering
    \includegraphics[width=\linewidth]{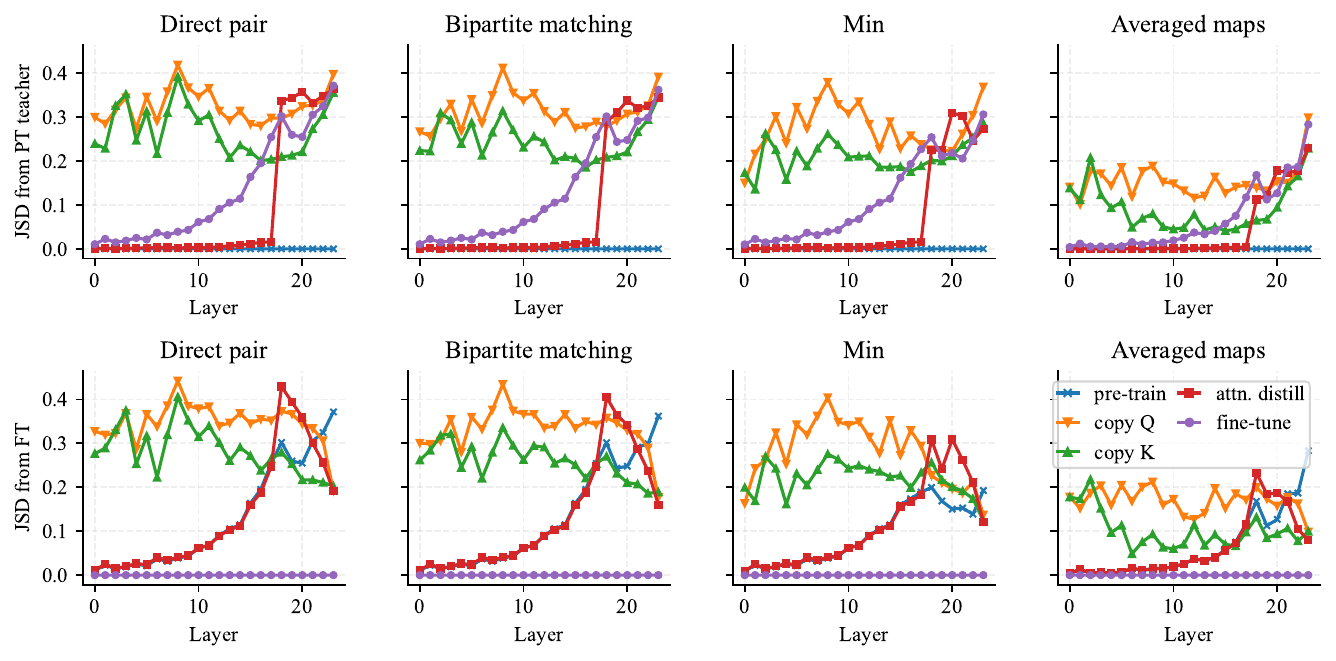}
    \caption{\textbf{Attention map similarity across methods}. Each column corresponds to a different way of matching up attention heads between two models. The top row shows the Jensen-Shannon divergence (JSD) with respect to the MAE pre-trained teacher, whereas the bottom row shows the JSD with respect to the fine-tuned MAE model. These plots match our intuition on distillation or fine-tuning methods, but copy $Q$ is consistently dissimilar from the PT and FT models. Note that this may be a limitation of this particular analysis, since there are many settings of $Q$, $K$, and $V$ that lead to the same layer output.}
    \label{fig:head_jsd}
\end{figure}

In \cref{sec:variants}, we found that copying the queries $Q$ does surprisingly well, almost matching \adistill or fine-tuning the pre-trained weights. Here, we compare the attention maps learned by the copy $Q$ model to those of other models, in hopes of understanding why copy $Q$ does so well. 

Each of the 24 layers within a ViT-L has 16 attention heads, which each compute an $L \times L$ attention map for an image with $L$ patches. We would like to determine the similarity between the attention heads in two models using some divergence measure; we use the Jensen-Shannon divergence (JSD) because it is symmetric in its arguments. However, there is one caveat. Because the output of the attention layer is invariant to the ordering of its heads, it is insufficient to compare the $i$th head of one model against the $i$th head of another. We need to properly match heads up across models. We explored four ways of doing so: 

\begin{enumerate}
    \item Direct pair: this is the naive approach of computing the JSD between the $i$th head of the first model and the $i$th head of the second model. This can fail since similar heads may not be in the same order across models. 
    \item Bipartite matching: for each layer, we compute the JSD between each of the 16 heads in the first model and the 16 heads in the second model. We then use bipartite matching to create a one-to-one pairing between the heads that minimizes the cumulative JSD. This solves the previous problem, but can still be thrown off, such as if one of the models has heads that it applies no weight to ($V=0$ or $W_{proj}=0$ or $W_{proj}$ is orthogonal to the values). 
    \item Minimum: instead of creating a one-to-one matching, we allow many-to-one matching between heads. We call this Minimum because each head in the first model is paired with the head from the second model with the smallest JSD. This allows our metric to potentially ignore extraneous heads in the second model, but is still susceptible to extraneous heads in the first model. 
    \item Averaged maps: we average the attention maps of all heads in a layer and compare the averaged maps across models. This can still be thrown off by extraneous heads. 
\end{enumerate}

\cref{fig:head_jsd} shows the results of comparing models against the pre-trained teacher (top row) or fine-tuned model (bottom row) as the second model in the JSD. Most of our findings align with our intuition. In the top row, when comparing against the pre-trained teacher, attention distillation matches the teacher maps closely until layer 18, the last layer whose attention maps it is trained to approximate. The fine-tuned model's attention maps diverge more in later layers, since layerwise learning rate decay ensures that the earlier layers don't change much. However, copy $Q$ is only somewhat similar to the pre-trained teacher or the fine-tuned model, across all of our ways to measure attention map similarity. Furthermore, it is less similar than copy $K$ is, even though copy $K$ has much lower downstream performance than copy $Q$. 

Note that these plots have major limitations in what kinds of similarity they capture. With enough attention heads per layer, the same exact attention map can be partitioned differently across the heads between two models. Hypothetically, let's say that an attention layer wants to attend uniformly across all locations (\textit{i.e.,} perform average pooling), and that we have 3 models, each with 2 attention heads:
\begin{enumerate}
    \item Head 1 attends uniformly over all locations, head 2 attends arbitrarily over locations, and the second head's values are set to 0.
    \item Head 1 attends uniformly over the top half of the image, head 2 attends uniformly over the bottom half of the image, and both  use values $V/2$.
    \item Head 1 attends uniformly over the left half of the image, head 2 attends uniformly over the right half of the image, and both  use values $V/2$.
\end{enumerate}
All 3 heads compute the same exact attention operation, yet would register as highly dissimilar in the setup from \cref{fig:head_jsd}.
Overall, this experiment shows that copy $Q$'s behavior is highly complex, and its strong downstream performance is still not fully understand.

\begin{figure*}[t]
\centering
\vspace{-.5em}
\includegraphics[width=0.95\linewidth]{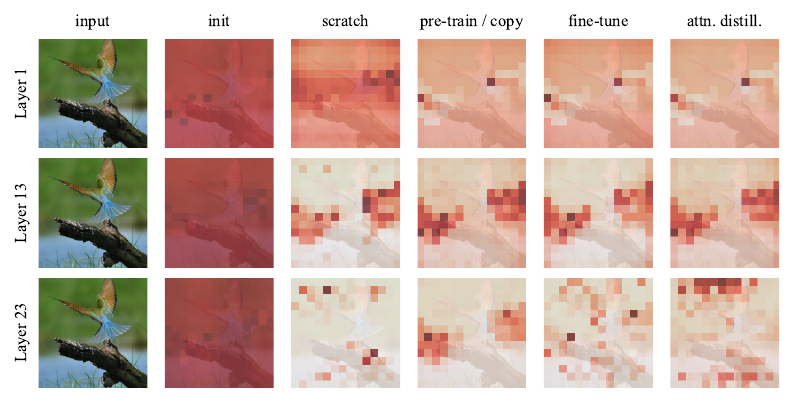}
\vspace{-1em}
\caption{\label{fig:attn_vis}\textbf{Visualization of attention maps for different methods}. We show what the [CLS] token attends to at various layers within the network. Darker patches indicate more attention weight. Notably, the pre-trained MAE model's attention maps provide a significant prior over what the model should use, separating the object from potentially spurious cues like the branch or the background. In contrast, models right at random initialization (``init'') start off attending uniformly over the image, which leads the scratch model to use more of the spurious patches. We show more attention map visualizations in Appendix~\ref{sec:more_attn_vis}.
}
\end{figure*}

\subsection{Attention Map Visualizations}
\label{sec:attn_map_vis}
\begin{figure}[!h]
    \centering
    \includegraphics[width=\linewidth]{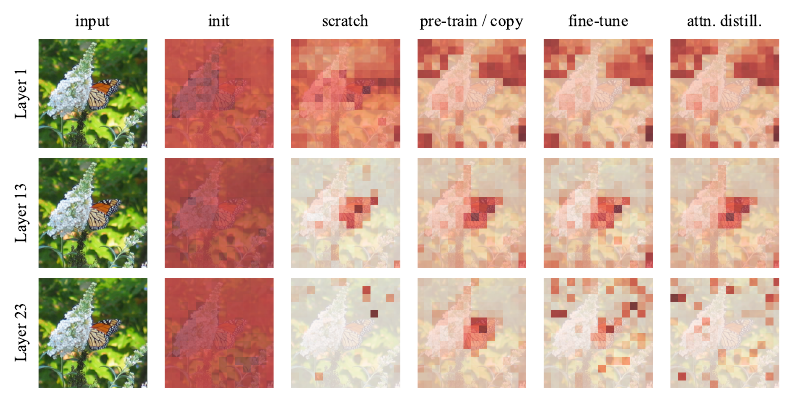} \\
    \includegraphics[width=\linewidth]{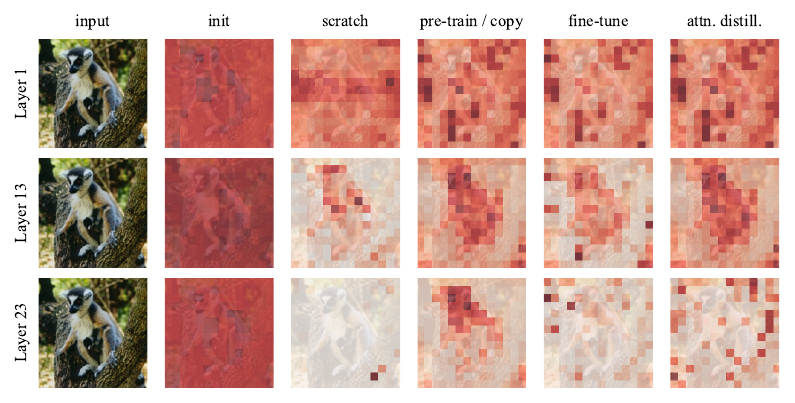}
    \caption{\textbf{Attention map visualizations on more examples}}
    \label{fig:more_attn_vis}
\end{figure}

\begin{figure}[!h]
    \centering
    \includegraphics[width=\linewidth]{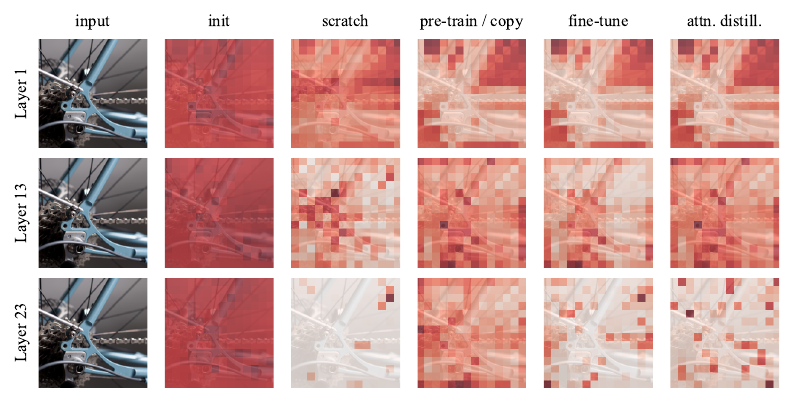} \\
    \includegraphics[width=\linewidth]{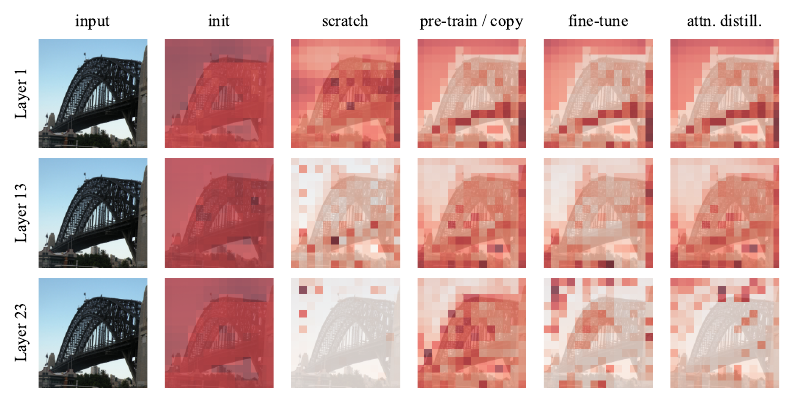} 
    \caption{\textbf{Attention map visualizations on more examples}}
    \label{fig:more_attn_vis2}
\end{figure}

\cref{sec:relearn} provided several results to show that attention transfer does not simply ``relearn'' the teacher. Here, we examine one final piece of evidence. We show the attention maps of different networks in \cref{fig:attn_vis}. We focus on \adistill, since \acopy's maps are identical to those in the teacher. Attention Distillation's maps generally match the teacher (pre-trained MAE), but are not completely identical for layers that are distilled (\textit{e.g.,} layer 13). For layer 24, which is not distilled, Attention Distillation looks very different from pre-trained model, instead resembling the attention map of a model trained from scratch. These visualizations also highlight the fact that these attention maps are a very strong prior on what the model should use. While the randomly initialized model attends completely uniformly over all tokens, the pre-trained teacher attention maps already separate the relevant object from potentially spurious correlations (like the branch or background in the example). We show additional attention map visualizations in \cref{fig:more_attn_vis} and \cref{fig:more_attn_vis2}.

\subsection{Attention Distillation Hyperparameter Sensitivity}
We show the sensitivity of Attention Distillation to its hyperparameters. 

\paragraph{Distillation loss weight} We first consider the distillation loss weight $\lambda$ which is used to compute the overall loss for the student: 
\begin{align}
    \mathcal L = \mathcal L_\text{task} + \lambda L_\text{dist}
\end{align}
Table~\ref{tab:distill_loss_weight} shows that a larger weight, $\lambda = 3$, does best. This may be because it encourages the student to learn useful attention maps more quickly, letting it guide feature learning earlier in training. We use this value of $\lambda$ for our main result, where we match the 85.7 accuracy of fine-tuning. However, all other results in this paper use $\lambda = 1$ for simplicity.

\paragraph{Partial distillation: layers} Just as we tried for Attention Copy, we also tried distilling various numbers of layers from the MAE teacher network, starting from the bottom of the network. Table~\ref{tab:distill_teacher_layers} shows that there is a ``sweet spot'' when distilling the first 21 out of 24 layers. Distilling all layers may hurt performance by forcing the student to use attention maps that are more suited for reconstruction than classification. Note that all other distillation results in this paper use the first 18 layers by default.

\begin{figure}[t]
    \begin{minipage}[b]{0.48\textwidth}
    \centering
    \tablestyle{8pt}{1.2}
    \begin{tabular}{y{90}x{60}}
        Distillation loss weight $\lambda$ & Accuracy \\
        \shline
        0.3 & 84.7 \\
        1.0 & 85.3 \\
        3.0 & 85.7 \\
    \end{tabular}
    \caption{\textbf{Weight on distillation loss}}
    \label{tab:distill_loss_weight}
    \end{minipage}
    \hfill
    \begin{minipage}[b]{0.48\textwidth}
    \centering
    \tablestyle{8pt}{1.2}
    \begin{tabular}{y{90}x{60}}
        Layers distilled & Accuracy \\
        \shline
        18 & 85.3 \\
        21 & 85.5 \\
        24 & 85.1 \\
    \end{tabular}
    \caption{\textbf{Number of layers distilled}}
    \label{tab:distill_teacher_layers}
    \end{minipage}
\end{figure}

\begin{table}
    \centering
    \tablestyle{8pt}{1.2}
    \begin{tabular}{y{40}x{35}x{90}}
        Teacher & Student & Accuracy \\
        \shline
        Random  & Random & 72.7 \\
        MAE     & Random & 85.1 \\
        MoCo-v3 & Random & 82.5 \\
        FLIP    & Random & 86.6 \\
        FLIP    & MAE    & 84.2 \\
        MAE     & MAE    & 85.4 \\
        MoCo-v3 & MAE    & 82.9 \\
        MAE     & FLIP   & 83.2 \\
    \end{tabular}
    \vspace{0.5em}
    \caption{\textbf{Decoupling attention maps from features}. We try Attention Distillation with various pre-trained students. }
    \label{tab:decouple}
\end{table}
\subsection{Mix and Match, Student and Teacher\label{sec:mix_match}}
In the main paper, we focused on transferring attention maps from a pre-trained teacher to a randomly initialized student. However, the fact that Transformers have decoupled inter- and intra-token computation means that we can actually initialize the student with a pre-trained network as well. This entails testing whether the attention patterns from one network can improve the features of an already-pre-trained student model. We try Attention Distillation for various combinations of MAE, MoCo-v3, FLIP, and a randomly initialized network. Table~\ref{tab:decouple} shows that this ``mix-and-match'' training does better than training from scratch (83.0) but does not match the performance in \cref{tab:init}, where the students are randomly initialized. These are preliminary results, as the overall training recipe may need to be changed to accommodate the different learning dynamics of a different student model. Further hyperparameter tuning may significantly improve these results.

\section{Implementation Details\label{sec:impl_details}}
We present the training recipe for \acopy in \cref{tab:copy_recipe} and the recipe for \adistill in \cref{tab:distill_recipe}. For our partial layer transfer experiments in \cref{fig:block}, we set $\beta_2 = 0.95$ as it helps avoid training instabilities. 

\begin{table}[h]
    \centering
    \begin{tabular}{l|l}
        Config & Value \\
        \midrule
        optimizer & AdamW~\cite{Loshchilov2019} \\
        base learning rate & 1e-3 \\
        minimum absolute lr & 2e-3  \\ 
        weight decay & 0.05 \\
        optimizer momentum & $\beta_1=0.9$, $\beta_2 = 0.999$ \\ 
        layerwise lr decay & 0.75 \\
        batch size & 2048 \\
        learning rate schedule & cosine decay~\cite{Loshchilov2016} \\
        warmup epochs~\cite{Goyal2017} & 5 \\
        training epochs & 100 \\
        augmentation & RandAug (9, 0.5)~\cite{Cubuk2020} \\
        label smoothing~\cite{Szegedy2016a} & 0.1 \\
        mixup~\cite{Zhang2018a} & 0.8 \\
        cutmix~\cite{Yun2019} & 1.0 \\
        drop path~\cite{Huang2016} & 0 \\
        exp. moving average (EMA) & 0.9999 \\
        layers copied & 24 \\
    \end{tabular}
    \vspace{0.5em}
    \caption{\textbf{Training recipe for Attention Copy on ViT-L}. }
    \label{tab:copy_recipe}
\end{table}

\begin{table}[h]
    \centering
    \begin{tabular}{l|l}
        Config & Value \\
        \midrule
        optimizer & AdamW \\
        base learning rate & 1e-4 \\
        weight decay & 0.3 \\
        optimizer momentum & $\beta_1=0.9$, $\beta_2 = 0.95$~\cite{Chen2020c} \\ 
        batch size & 2048 \\
        learning rate schedule & cosine decay \\
        warmup epochs & 20 \\
        training epochs & 200 \\
        augmentation & RandAug (9, 0.5) \\
        label smoothing & 0.1 \\
        mixup & 0.8 \\
        cutmix & 1.0 \\
        drop path & 0.2 \\
        exp. moving average (EMA) & 0.9999 \\
        layers copied & 18 \\
        distillation weight $\lambda$ & 3
    \end{tabular}
    \vspace{0.5em}
    \caption{\textbf{Training recipe for Attention Distillation on ViT-L}. }
    \label{tab:distill_recipe}
\end{table}

\clearpage
\newpage
\section{Additional Attention Map Visualizations}
\label{sec:more_attn_vis}

\end{document}